\documentclass[11pt,a4paper]{article}
\usepackage{url}
\usepackage[hyperindex,breaklinks]{hyperref}
\usepackage{breakurl}

\usepackage[hyperref]{acl2017}
\usepackage{times}
\usepackage{latexsym}

\usepackage{amssymb}
\usepackage{amsthm}
\usepackage{amsmath}
\usepackage{mathrsfs}
\usepackage{adjustbox}
\usepackage{natbib}
\usepackage{epsfig}
\usepackage{booktabs}

\usepackage{stix}
\usepackage{enumitem}

\usepackage{caption}
\usepackage{multirow}
\usepackage{placeins}
\usepackage{enumitem}
\setlist{noitemsep} 

\aclfinalcopy 
\def\aclpaperid{3129} 

\newsavebox\CBox
\def\textBF#1{\sbox\CBox{#1}\resizebox{\wd\CBox}{\ht\CBox}{\textbf{#1}}}

\title{Character Composition Model with Convolutional Neural Networks for Dependency Parsing on Morphologically Rich Languages}

\author{Xiang Yu \and Ngoc Thang Vu \\
  Institut f\"ur Maschinelle Sprachverarbeitung\\
  Universit\"at Stuttgart\\
  {\tt \{xiangyu,thangvu\}@ims.uni-stuttgart.de} }

\date{}

\begin{document}
\maketitle
\begin{abstract}
  We present a transition-based dependency parser that uses a convolutional neural network to compose word representations from characters. The character composition model shows great improvement over the word-lookup model, especially for parsing agglutinative languages. These improvements are even better than using pre-trained word embeddings from extra data. On the SPMRL data sets, our system outperforms the previous best greedy parser \citep{Ballesteros:2015} by a margin of 3\% on average.\footnote{The parser is available at \burl{http://www.ims.uni-stuttgart.de/institut/mitarbeiter/xiangyu/index.en.html}}

\end{abstract}

\section{Introduction}
  

  As with many other NLP tasks, dependency parsing also suffers from the out-of-vocabulary (OOV) problem, and probably more than others since training data with syntactical annotation is usually scarce.
  This problem is particularly severe when the target is a morphologically rich language. For example, in the SPMRL shared task data sets \citep{Seddah:2013, Seddah:2014}, 4 out of 9 treebanks contain more than 40\% word types in the development set that are never seen in the training set.

  One way to tackle the OOV problem is to pre-train the word embeddings, {\em e.g.}, with word2vec \citep{Mikolov:2013}, from a large set of unlabeled data. 
  This comes with two main advantages: (1) more word types, which means that the vocabulary is extended by the unlabeled data, so that some of the OOV words now have a learned representation; (2) more word tokens per type, which means that the syntactic and semantic similarities of the words are better modeled than only using the parser training data.

  Pre-trained word embeddings can alleviate the OOV problem by expanding the vocabulary,
  but it does not model the morphological information.
  Instead of looking up word embeddings, many researchers propose to compose the word representation from characters for various tasks, {\em e.g.}, part-of-speech tagging \citep{Santos:2014, Plank:2016}, named entity recognition \citep{Santos:2015}, language modeling \citep{Ling:2015}, machine translation \citep{Costa:2016}.
  In particular, \citet{Ballesteros:2015} use a bidirectional long short-term memory (LSTM) character model for dependency parsing. \citet{Kim:2016} present a convolutional neural network (CNN) character model for language modeling, but make no comparison among the character models, and state that ``it remains open as to which character composition model ({\em i.e.}, LSTM or CNN) performs better''.

  We propose to apply the CNN model by \citet{Kim:2016} in a greedy transition-based dependency parser with feed-forward neural networks \citep{Chen:2014, Weiss:2015}. This model requires no extra unlabeled data but performs better than using pre-trained word embeddings. Furthermore, it can be combined with word embeddings from the lookup table since they capture different aspects of word similarities.

  Experimental results show that the CNN model works especially well on agglutinative languages, where the OOV rates are high. On other morphologically rich languages, the CNN model also performs at least as good as the word-lookup model.

  Furthermore, our CNN model outperforms both the original and our re-implementation of the bidirectional LSTM model by \citet{Ballesteros:2015} by a large margin. It provides empirical evidence to the aforementioned open question, suggesting that the CNN is the better character composition model for dependency parsing.



\section{Parsing Models}

  \subsection{Baseline Parsing Model}\label{sec:baseline}
    As the baseline parsing model, we re-implement the greedy parser in \citet{Weiss:2015} with some modifications, which brings about 0.5\% improvement, outlined below.\footnote{We only experiment with the greedy parser, since this paper focuses on the character-level input features and is independent of the global training and inference as in \citet{Weiss:2015, Andor:2016}.}

    Since most treebanks contain non-projective trees, we use an approximate non-projective transition system similar to \citet{Attardi:2006}. It has two additional transitions (\textsc{Left-2} and \textsc{Right-2}) to the Arc-Standard system \citep{Nivre:2004} that attach the top of the stack to the third token on the stack, or {\em vice versa}. 
    We also extend the feature templates in \citet{Weiss:2015} by extracting the children of the third token in the stack. The complete transitions and feature templates are listed in Appendix~\ref{sec:system}.

    The feature templates consist of 24 tokens in the configuration, we look up the embeddings of the word forms, POS tags and dependency labels of each token.\footnote{The tokens in the stack and buffer do not have labels yet, we use a special label \texttt{<NOLABEL>} instead.} We then concatenate the embeddings $\mathbf{E}_{word}(t_i)$, $\mathbf{E}_{tag}(t_i)$, $\mathbf{E}_{label}(t_i)$ for each token $t_i$, and use a dense layer with ReLU non-linearity to obtain a compact representation $\mathbf{f}(t_i)$ of the token:
    \begin{align*}
    \mathbf{x}(t_i) &= [\mathbf{E}_{word}(t_i); \mathbf{E}_{tag}(t_i); \mathbf{E}_{label}(t_i)] \tag{1}\\
    \mathbf{f}(t_i) &= \max\{\mathbf{0}, \mathbf{W}_f \mathbf{x}(t_i) + \mathbf{b}_f\}
    \end{align*}

    We concatenate the representations of the tokens and feed them through two hidden layers with ReLU non-linearity, and finally into the softmax layer to compute the probability of each transition:
    \begin{align*}
      \mathbf{h}_0 &= [\mathbf{f}(t_1); \mathbf{f}(t_2); ... ; \mathbf{f}(t_{24})] \\
      \mathbf{h}_1 &= \max\{\mathbf{0}, \mathbf{W}_1 \mathbf{h}_0 + \mathbf{b}_1\} \\
      \mathbf{h}_2 &= \max\{\mathbf{0}, \mathbf{W}_2 \mathbf{h}_1 + \mathbf{b}_2\} \\
      p(\cdot | t_1, ..., t_{24}) &= \mathrm{softmax}(\mathbf{W}_3 \mathbf{h}_2 + \mathbf{b}_3)
    \end{align*}

    $\mathbf{E}_{word}$, $\mathbf{E}_{tag}$, $\mathbf{E}_{label}$, $\mathbf{W}_f$, $\mathbf{W}_1$, $\mathbf{W}_2$, $\mathbf{W}_3$, $\mathbf{b}_f$, $\mathbf{b}_1$, $\mathbf{b}_2$, $\mathbf{b}_3$ are all the parameters that will be learned through back propagation with averaged stochastic gradient descent in mini-batches.

    Note that \citet{Weiss:2015} directly concatenate the embeddings of the words, tags, and labels of all the tokens together as input to the hidden layer. Instead, we first group the embeddings of the word, tag, and label of each token and compute an intermediate representation with shared parameters, then concatenate all the representations as input to the hidden layer. 

  \subsection{LSTM Character Composition Model}
    To tackle the OOV problem, we want to replace the word-lookup table with a function that composes the word representation from characters. 

    As a baseline character model, we re-implement the bidirectional LSTM character composition model following \citet{Ballesteros:2015}. We replace the lookup table $\mathbf{E}_{word}$ in the baseline parser with the final outputs of the forward and backward LSTMs $\overleftarrow{\mathbf{lstm}}$ and $\overrightarrow{\mathbf{lstm}}$.
    Equation (1) is then replaced with
    \begin{equation*}
    \mathbf{x}(t_i) = [\overleftarrow{\mathbf{lstm}}(t_i); \overrightarrow{\mathbf{lstm}}(t_i); \mathbf{E}_{tag}(t_i); \mathbf{E}_{label}(t_i)] .\\
    \end{equation*} 
    We refer the readers to \citet{Ling:2015} for the details of the bidirectional LSTM.

  \subsection{CNN Character Composition Model}
    In contrast to the LSTM model, we propose to use a ``flat'' CNN as the character composition model, similar to \citet{Kim:2016}.\footnote{We do not use the highway networks since it did not improve the performance in preliminary experiments.}

    Equation~(1) is thus replaced with
    \begin{align*}
    \mathbf{x}(t_i) = & [\mathbf{cnn}^{l_1}(t_i); \mathbf{cnn}^{l_2}(t_i); ...; \mathbf{cnn}^{l_k}(t_i); \\
           & \mathbf{E}_{tag}(t_i); \mathbf{E}_{label}(t_i)] .\tag{2}
    \end{align*} 
    
    Concretely, the input of the model is a concatenated matrix of character embeddings $\mathbf{C} \in \mathbb{R}^{d_i \times w}$, where $d_i$ is the dimensionality of character embeddings (number of input channels) and $w$ is the length of the padded word.\footnote{The details of the padding is in Appendix~\ref{sec:preprocess}.} 
    We apply $k$ convolutional kernels $\mathcal{K} \in \mathbb{R}^{d_o \times d_i \times l_k}$ with ReLU non-linearity on the input, where $d_o$ is the number of output channels and $l_k$ is the length of the kernel.
    The output of the convolution operation is $\mathbf{O}_{conv} \in \mathbb{R}^{d_o \times (l-k+1)}$, and we apply a max-over-time pooling that takes the maximum activations of the kernel along each channel, obtaining the final output $\mathbf{O}_{final} \in \mathbb{R}^{d_o}$, which corresponds to the most salient n-gram representation of the word, denoted $\mathbf{cnn}^{l_k}$ in Equation~(2). 
    We then concatenate the outputs of several such CNNs with different lengths, so that the information from different n-grams are extracted and can interact with each other. 




    \begin{table*}[!h]
    \small
    \centering
    \begin{tabular}{r | r | r r r r r r r r r | r }
    \multicolumn{2}{c|}{Model}  & Ara  & Baq & Fre & Ger & Heb & Hun & Kor & Pol & Swe & Avg \\
    \noalign{\hrule height 0.5pt}
    \multirow{8}{*}{\texttt{Int}} & \texttt{WORD}      & 84.50 & 77.87 & 82.20 & 85.35 & \textBF{74.68} & 76.17 & 84.62 & 80.71 & 79.14 & 80.58 \\
    &\texttt{W2V}       & \textBF{85.11} & 79.07 & \textBF{82.73} & \textBF{86.60} & 74.55 & 78.21 & 85.30 & 82.37 & 79.67 & 81.51 \\
    &\texttt{LSTM}      & 83.42 & 82.97 & 81.35 & 85.34 & 74.03 & 83.06 & 86.56 & 80.13 & 77.44 & 81.48 \\
    &\texttt{CNN}       & 84.65 & \textBF{83.91} & 82.41 & 85.61 & 74.23 & \textBF{83.68} & \textBF{86.99} & \textBF{83.28} & \textBF{80.00} & \textBF{82.75} \\
    \cline{2-12}
    &\texttt{LSTM+WORD} & \textBF{84.75} & 83.43 & \textBF{82.25} & 85.56 & 74.62 & 83.43 & 86.85 & 82.30 & 79.85 & 82.56 \\
    &\texttt{CNN+WORD}  & 84.58 & \textBF{84.22} & 81.79 & \textBF{85.85} & \textBF{74.79} & \textBF{83.51} & \textBF{87.21} & \textBF{83.66} & \textBF{80.52} & \textBF{82.90} \\
    \cline{2-12}
    &\texttt{LSTM+W2V}   & 85.35 & 83.94 & 83.04 & 86.38 & \textBF{75.15} & 83.30 & 87.35 & 83.00 & 79.38 & 82.99 \\
    &\texttt{CNN+W2V}   & \textBF{85.67} & \textBF{84.37} & \textBF{83.09} & \textBF{86.81} & 74.95 & \textBF{84.08} & \textBF{87.72} & \textBF{84.44} & \textBF{80.35} & \textBF{83.50} \\
    \noalign{\hrule height 0.5pt}
    \multirow{3}{*}{\texttt{Ext}} &\texttt{B15-WORD} & \textBF{83.46} & 73.56 & \textBF{82.03} & \textBF{84.62} & \textBF{72.70} & 69.31 & 83.37 & \textBF{79.83} & \textBF{76.40} & 78.36 \\
    &\texttt{B15-LSTM} & 83.40 & \textBF{78.61} & 81.08 & 84.49 & 72.26 & \textBF{76.34} & \textBF{86.21} & 78.24 & 74.47 & \textBF{79.46}\\
    \cline{2-12}
    &\texttt{BestPub}      & 86.21 & 85.70 & 85.66 & 89.65 & 81.65 & 86.13 & 87.27 & 87.07 & 82.75 & 85.79\\
    \end{tabular}
    \caption{LAS on the test sets, the best LAS in each group is marked in bold face.}\label{tab:results}
    \end{table*}

\section{Experiments}

  \subsection{Experimental Setup}
    We conduct our experiments on the treebanks from the SPMRL 2014 shared task \citep{Seddah:2013, Seddah:2014}, which includes 9 morphologically rich languages: Arabic, Basque, French, German, Hebrew, Hungarian, Korean, Polish, and Swedish.
    All the treebanks are split into training, development, and test sets by the shared task organizers.
    We use the fine-grained predicted POS tags provided by the organizers, and evaluate the labeled attachment scores (LAS) including punctuation. 

    We experiment with the CNN-based character composition model (\texttt{CNN}) along with several baselines. The first baseline (\texttt{WORD}) uses the word-lookup model described in Section~\ref{sec:baseline} with randomly initialized word embeddings. The second baseline (\texttt{W2V}) uses pre-trained word embeddings by word2vec \citep{Mikolov:2013} with the CBOW model and default parameters on the unlabeled texts from the shared task organizers. The third baseline (\texttt{LSTM}) uses a bidirectional LSTM as the character composition model following \citet{Ballesteros:2015}.
    Appendix~\ref{sec:hyper} lists the hyper-parameters of all the models.

    Further analysis suggests that combining the character composition models with word-lookup models could be beneficial since they capture different aspects of word similarities (orthographic vs. syntactic/semantic). We therefore experiment with four combined models in two groups: (1) randomly initialized word embeddings (\texttt{LSTM+WORD} vs. \texttt{CNN+WORD}), and (2) pre-trained word embeddings (\texttt{LSTM+W2V} vs. \texttt{CNN+W2V}).

    The experimental results are shown in Table~\ref{tab:results}, with \texttt{Int} denoting internal comparisons (with three groups) and \texttt{Ext} denoting external comparisons, the highest LAS in each group is marked in bold face.

  \subsection{Internal Comparisons}
    In the first group, we compare the LAS of the four single models \texttt{WORD}, \texttt{W2V}, \texttt{LSTM}, and \texttt{CNN}.
    In macro average of all languages, the \texttt{CNN} model performs 2.17\% higher than the \texttt{WORD} model, and 1.24\% higher than the \texttt{W2V} model. 
    The \texttt{LSTM} model, however, performs only 0.9\% higher than the \texttt{WORD} model and 1.27\% lower than the \texttt{CNN} model. 

    The CNN model shows large improvement in four languages: three agglutinative languages (Basque, Hungarian, Korean), and one highly inflected fusional language (Polish). They all have high OOV rate, thus difficult for the baseline parser that does not model morphological information. Also, morphemes in agglutinative languages tend to have unique, unambiguous meanings, thus easier for the convolutional kernels to capture.

    In the second group, we observe that the additional word-lookup model does not significantly improve the CNN moodel (from 82.75\% in \texttt{CNN} to 82.90\% in \texttt{CNN+WORD} on average) while the LSTM model is improved by a much larger margin (from 81.48\% in \texttt{LSTM} to 82.56\% in \texttt{LSTM+WORD} on average). This suggests that the CNN model has already learned the most important information from the the word forms, while the LSTM model has not.
    Also, the combined \texttt{CNN+WORD} model is still better than the \texttt{LSTM+WORD} model, despite the large improvement in the latter.

    In the third group where pre-trained word embeddings are used, combining \texttt{CNN} with \texttt{W2V} brings an extra 0.75\% LAS over the \texttt{CNN} model. Combining \texttt{LSTM} with \texttt{W2V} shows similar trend but with much larger improvement, yet again, \texttt{CNN+W2V} outperforms \texttt{LSTM+W2V}
    both on average and individually in 8 out of 9 languages.




    \begin{table*}[t!]
    \small
    \centering
    \begin{tabular}{ r | r | r r r r r r r r r | r }
    Model & Case  & Ara  & Baq & Fre & Ger & Heb & Hun & Kor & Pol & Swe & Avg \\
    \noalign{\hrule height 0.5pt}
    \multirow{2}{*}{\texttt{CNN}} & \texttt{$\Delta$IV} & 0.12  & 2.72  & -0.44 & 0.13  & -0.35 & 1.48  & 1.30  & 0.98  & 1.39  & 0.81  \\
    & \texttt{$\Delta$OOV}  & 0.03  & 5.78  & 0.33  & 0.10  & -1.04 & 5.04  & 2.17  & 2.34  & 0.95  & 1.74  \\
    \hline
    \multirow{2}{*}{\texttt{LSTM}} & \texttt{$\Delta$IV} & -0.58 & 1.98  & -0.55 & -0.08 & -1.23 & 1.62  & 1.12  & -0.49 & 0.21  & 0.22  \\
    & \texttt{$\Delta$OOV}  & -0.32 & 5.09  & 0.12  & -0.21 & -1.99 & 4.74  & 1.51  & 0.10  & 0.38  & 1.05  \\
    \end{tabular}
    \caption{LAS improvements by \texttt{CNN} and \texttt{LSTM} in the IV and OOV cases on the development sets.} \label{tab:oov}
    \end{table*}

      \begin{table*}[h!]
    \small
    \centering
    \begin{tabular}{  r | r r r r r r r r r | r }
    Mod  & Ara  & Baq & Fre & Ger & Heb & Hun & Kor & Pol & Swe & Avg \\
    \noalign{\hrule height 0.5pt}
    $\clubsuit$\texttt{b}\texttt{c}  & -1.23 & -1.94 & -1.35 & -1.57 & -0.79 & -3.23 & -1.22 & -2.53 & -1.54 & -1.71 \\
    \texttt{a}$\clubsuit$\texttt{c}  & -3.47 & -3.96 & -2.39 & -2.54 & -1.24 & -4.52 & -3.21 & -4.47 & -4.19 & -3.33 \\
    \texttt{a}\texttt{b}$\clubsuit$  & -1.52 & -15.31  & -0.72 & -1.23 & -0.26 & -13.97  & -10.22  & -3.52 & -2.61 & -5.48 \\
    \noalign{\hrule height 0.5pt}
    \texttt{a}$\clubsuit$$\clubsuit$ & -3.73 & -19.29  & -3.30 & -3.49 & -1.21 & -17.89  & -12.95  & -6.22 & -6.01 & -8.23 \\
    $\clubsuit$\texttt{b}$\clubsuit$ & -3.02 & -18.06  & -2.60 & -3.54 & -1.42 & -18.43  & -11.69  & -6.22 & -3.85 & -7.65 \\
    $\clubsuit$$\clubsuit$\texttt{c} & -5.11 & -7.90 & -4.05 & -4.86 & -2.50 & -9.75 & -4.56 & -6.71 & -6.74 & -5.80 \\
    \end{tabular}
    \caption{Degradation of LAS of the \texttt{CNN} model on the masked development sets.}\label{tab:mod}
    \end{table*}

  \subsection{External Comparisons}

    We also report the results of the two models from \citet{Ballesteros:2015}: \texttt{B15-WORD} with randomly initialized word embeddings and \texttt{B15-LSTM} as their proposed model. 
    Finally, we report the best published results (\texttt{BestPub}) on this data set \citep{Bjorkelund:2013, Bjorkelund:2014}. 

    On average, the \texttt{B15-LSTM} model improves their own baseline by 1.1\%, similar to the 0.9\% improvement of our \texttt{LSTM} model, which is much smaller than the 2.17\% improvement of the \texttt{CNN} model.
    Furthermore, the \texttt{CNN} model is improved from a strong baseline: our \texttt{WORD} model performs already 2.22\% higher than the \texttt{B15-WORD} model.

    Comparing the individual performances on each language, we observe that the \texttt{CNN} model almost always outperforms the \texttt{WORD} model except for Hebrew.
    However, both \texttt{LSTM} and \texttt{B15-LSTM} perform higher than baseline only on the three agglutinative languages (Basque, Hungarian, and Korean), and lower than baseline on the other six. 

    \citet{Ballesteros:2015} do not compare the effect of adding a word-lookup model to the LSTM model as in our second group of internal comparisons. However, \citet{Plank:2016} show that combining the same LSTM character composition model with word-lookup model improves the performance of POS tagging by a very large margin. This partially confirms our hypothesis that the LSTM model does not learn sufficient information from the word forms.

    Considering both internal and external comparisons in both average and individual performances, we argue that CNN is more suitable than LSTM as character composition model for parsing.

    While comparing to the best published results \citep{Bjorkelund:2013, Bjorkelund:2014}, we have to note that their approach uses explicit morphological features, ensemble, ranking, 
    {\em etc.}, which all can boost parsing performance. 
    We only use a greedy parser with much fewer features, but bridge the 6 points gap between the previous best greedy parser and the best published result by more than one half.


  \subsection{Discussion on CNN and LSTM}

    We conjecture that the main reason for the better performance of CNN over LSTM is its flexibility in processing sub-word information.
    The CNN model uses different kernels to capture n-grams of different lengths. In our setting, a kernel with a minimum length of 3 can capture short morphemes; and with a maximum length of 9, it can practically capture a normal word. 
    With the flexibility of capturing patterns from morphemes up to words, the CNN model almost always outperforms the word-lookup model.

    In theory, LSTM has the ability to model much longer sequences, however, it is composed step by step with recurrence. For such deep network architectures, more data would be required to learn the same sequence, in comparison to CNN which can directly use a large kernel to match the pattern. 
    For dependency parsing, training data is usually scarce, this could be the reason that the LSTM has not utilized its full potential. 

    

\subsection{Analyses on OOV and Morphology}
    The motivation for using character composition models is based on the hypothesis that it can address the OOV problem.
    To verify the hypothesis, we analyze the LAS improvements of the \texttt{CNN} and \texttt{LSTM} model on the development sets in two cases: (1) both the head and the dependent are in vocabulary or (2) at least one of them is out of vocabulary. 
    Table~\ref{tab:oov} shows the results, where the two cases are denoted as \texttt{$\Delta$IV} and  \texttt{$\Delta$OOV}.
    The general trend in the results is that the improvements of both models in the OOV case are larger than in the IV case, which means that the character composition models indeed alleviates the OOV problem. 
    Also, \texttt{CNN} improves on seven languages in the IV case and eight languages in the OOV case, and it performs consistently better than \texttt{LSTM} in both cases.

    To analyze the informativeness of the morphemes at different positions, we conduct an ablation experiment.
    We split each word equally into three thirds, approximating the prefix, stem, and suffix.
    Based on that, we construct six modified versions of the development sets, in which we mask one or two third(s) of the characters in each word.
    Then we parse them with the \texttt{CNN} models trained on normal data. Table~\ref{tab:mod} shows the degradations of LAS on the six modified data sets compared to parsing the original data, where the position of $\clubsuit$ signifies the location of the masks.
    The three agglutinative languages Basque, Hungarian, and Korean suffer the most with masked words. In particular, the suffixes are the most informative for parsing in these three languages, since they cause the most loss while masked, and the least loss while unmasked. 
    The pattern is quite different on the other languages, in which the distinction of informativeness among the three parts is much smaller.

\section{Conclusion}
  In this paper, we propose to use a CNN to compose word representations from characters for dependency parsing. Experiments show that the CNN model consistently improves the parsing accuracy, especially for agglutinative languages. 
  In an external comparison on the SPMRL data sets, our system outperforms the previous best greedy parser.

  We also provide empirical evidence and analysis, showing that the CNN model indeed alleviates the OOV problem and that
  it is better suited than the LSTM in dependency parsing.

\section*{Acknowledgements}
  This work was supported by the German Research Foundation (DFG) in project D8 of SFB 732. We also thank our collegues in the IMS, especially Anders Bj\"orkelund, for valuable discussions, and the anonymous reviewers for the suggestions.


\FloatBarrier

\bibliography{acl2017}
\bibliographystyle{acl_natbib}

\newpage
\appendix
\section{Transitions and Feature Templates}\label{sec:system}

    \begin{table}[h!]
    \begin{tabular}{r | l }
      Transition & Configurations (Before $\Rightarrow$ After)\\
      \hline
      \textsc{Shift} &    $(\sigma, [w_i|\beta], A) \Rightarrow  ([\sigma|w_i], \beta, A) $  \\
      {\textsc{Left}} & $([\sigma|w_i, w_j], \beta, A)$\\
      & $\Rightarrow([\sigma|w_j], \beta, A\cup{(w_j \rightarrow w_i)}) $ \\
      {\textsc{Right}} & $([\sigma|w_i, w_j], \beta, A)$\\
      & $\Rightarrow([\sigma|w_i], \beta, A\cup{(w_i \rightarrow w_j)}) $ \\
      {\textsc{Left-2}} & $([\sigma|w_i, w_j, w_k], \beta, A)$\\
      & $\Rightarrow([\sigma|w_j, w_k], \beta, A\cup{(w_k \rightarrow w_i)}) $ \\
      {\textsc{Right-2}} & $([\sigma|w_i, w_j, w_k], \beta, A)$\\
      & $\Rightarrow([\sigma|w_i, w_j], \beta, A\cup{(w_i \rightarrow w_k)}) $ \\
    \end{tabular}
    \caption{The transition system in our experiments, where the configuration is a tuple of (stack, buffer, arcs).}\label{tab:feat}
    \end{table}


    \begin{table}[h!]
    \begin{tabular}{ l }
    \hline
    $s_1$, $s_2$, $s_3$, $s_4$, $b_1$, $b_2$, $b_3$, $b_4$, \\
    $s_1.lc_1$, $s_1.lc_2$, $s_1.rc_1$, $s_1.rc_2$, \\
    $s_2.lc_1$, $s_2.lc_2$, $s_2.rc_1$, $s_2.rc_2$, \\
    $s_3.lc_1$, $s_3.lc_2$, $s_3.rc_1$, $s_3.rc_2$, \\
    $s_1.lc_1.lc_1$, $s_1.lc_1.rc_1$, $s_1.rc_1.lc_1$, $s_1.rc_1.rc_1$\\
    \hline
    \end{tabular}
    \caption{The list of tokens to extract feature templates, where $s_i$ denotes the $i$-th token in the stack, $b_i$ the $i$-th token in the buffer, $lc_i$ denotes the $i$-th leftmost child, $rc_i$ the $i$-th rightmost child.}\label{tab:feat}
    \end{table}

\section{Character Input Preprocessing}\label{sec:preprocess}



  For the CNN input, we use a list of characters with fixed length to for batch processing.
  We add some special symbols apart from the normal alphabets, digits, and punctuations: \texttt{<SOW>} as the start of the word, \texttt{<EOW>} as the end of the word, \texttt{<MUL>} as multiple characters in the middle of the word squeezed into one symbol, \texttt{<PAD>} as padding equally on both sides, and \texttt{<UNK>} as characters unseen in the training data.

  For example, if we limit the input length to 9, a short word {\em ein} will be converted into \texttt{<PAD>}-\texttt{<PAD>}-\texttt{<SOW>}-e-i-n-\texttt{<EOW>}-\texttt{<PAD>}-\texttt{<PAD>}; a long word {\em pr\"achtiger} will be \texttt{<SOW>}-p-r-\"a-\texttt{<MUL>}-g-e-r-\texttt{<EOW>}. In practice, we set the length as 32, which is long enough for almost all the words.

\section{Hyper-Parameters}\label{sec:hyper}
  The common hyper-parameters of all the models are tuned on the development set in favor of the \texttt{WORD} model:
  \begin{itemize}[leftmargin=*]
    \item 100,000 training steps with random sampling of mini-batches of size 100;
    \item test on the development set every 2,000 steps;
    \item early stop if the LAS on the development does not improve for 3 times in a row;
    \item learning rate of 0.1, with exponential decay rate of 0.95 for every 2,000 steps;
    \item L2-regularization rate of $10^{-4}$;
    \item averaged SGD with momentum of 0.9;
    \item parameters are initialized following \citet{He:2015};
    \item dimensionality of the embeddings of each word, tag, and label are 256, 32, 32, respectively;
    \item dimensionality of the hidden layers are 512, 256;
    \item dropout on both hidden layers with rate of 0.1;
    \item total norm constraint of the gradients is 10.
  \end{itemize}

  The hyper-parameters for the \texttt{CNN} model are:
  \begin{itemize}[leftmargin=*]
    \item dimensionality of the character embedding is 32;
    \item 4 convolutional kernels of lengths 3, 5, 7, 9;
    \item number of output channels of each kernel is 64;
    \item fixed length for the character input is 32.
  \end{itemize}

  The hyper-parameters for the \texttt{LSTM} model are:
  \begin{itemize}[leftmargin=*]
    \item 128 hidden units for both LSTMs;
    \item all the gates use orthogonal initialization;
    \item gradient clipping of 10;
    \item no L2-regularization on the parameters.
  \end{itemize}

\end{document}